# Bridging the Accuracy Gap for 2-bit Quantized Neural Networks (QNN)

**Jungwook Choi** [1]  **Pierce I-Jen Chuang** [1]  **Zhuo Wang** [1,2]  **Swagath Venkataramani** [1]  **Vijayalakshmi Srinivasan** [1]
**Kailash Gopalakrishnan** [1]

## Abstract

Deep learning algorithms achieve high classification accuracy at the expense of significant computation cost. In order to reduce this cost, several quantization schemes have gained attention recently with some focusing on weight quantization, and others focusing on quantizing activations. This paper proposes novel techniques that target weight and activation quantizations separately resulting in an overall quantized neural network (QNN). The activation quantization technique, PArameterized Clipping acTivation (PACT), uses an activation clipping parameter $\alpha$ that is optimized during training to find the right quantization scale. The weight quantization scheme, statistics-aware weight binning (SAWB), finds the optimal scaling factor that minimizes the quantization error based on the statistical characteristics of the distribution of weights without the need for an exhaustive search. The combination of PACT and SAWB results in a 2-bit QNN that achieves state-of-the-art classification accuracy (comparable to full precision networks) across a range of popular models and datasets.

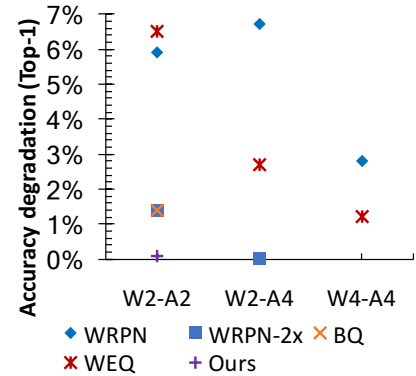

*Figure 1.* AlexNet: Accuracy degradation from previous QNN schemes and our 2-bit QNN = PACT + SAWB scheme. The proposed scheme achieves zero accuracy degradation for AlexNet quantized down to 2-bits for weights and activations without requiring any increase in the network size.

## 1. Introduction

Deep Convolutional Neural Networks (CNNs) have achieved remarkable accuracy for tasks in a wide range of application domains including image processing (He et al., 2016b), machine translation (Gehring et al., 2017), and speech recognition (Zhang et al., 2017). These state-of-the-art CNNs use very deep models, consuming 100s of ExaOps of computation during training and GBs of storage for model and data. This poses a tremendous challenge to widespread deployment, especially in resource constrained edge environments - leading to a plethora of explorations in compressed models that minimize memory footprint and computation while preserving model accuracy as much as possible.

Recently, a whole host of different techniques have been proposed to alleviate these computational costs. Among them, reducing the bit-precision of key CNN data structures, namely weights and activations, has gained attention due to its potential to significantly reduce both storage requirements and computational complexity. Several weight quantization techniques (Li & Liu, 2016; Zhu et al., 2017; Jan Achterhold, 2018; Antonio Polino, 2018; Lu Hou, 2018) have been proposed to reduce bit-precision of CNN weights but end up sacrificing model accuracy. Furthermore, a straightforward extension of weight quantization schemes to activation quantizations was also proposed, but it incurred significant accuracy degradation in large-scale image classification tasks such as IMAGENET (Russakovsky et al., 2015). Lately, activation quantization schemes based on greedy layer-wise optimization were proposed (Park et al., 2017; Zhou et al., 2017; Cai et al., 2017), but they required expensive processing (e.g., Lloyd's algorithm for activation quantization) in (Cai et al., 2017)), and achieved limited accuracy improvement.

Complementary to quantization schemes, increasing the network size has been shown to compensate accuracy loss due to quantization. For example, (Asit Mishra, 2018b) and (McDonnell, 2018) employed Wide Residual Network (Zagoruyko & Komodakis, 2016) for weight and activation quantization and demonstrated that increasing the number of channels reduces the quantization errors and allows more aggressive bit-width reductions. However, increasing the network size leads to a quadratic increase in the number of operations, which in turn increases the classification latency.

This work is motivated by the desire to significantly improve quantization schemes and achieve accuracy comparable to full-precision models while requiring no changes to the network structure - thereby harnessing the full computational benefits of quantization. We propose individual techniques targeting activation and weight quantizations resulting in an overall quantized neural network (QNN). The activation quantization technique, PArameterized Clipping acTivation (PACT), uses an activation clipping parameter $\alpha$ that is optimized during training to find the right quantization scale. The weight quantization scheme, statistics-aware weight binning (SAWB), finds the optimal scaling factor that minimizes the quantization error based on the statistical characteristics of the distribution of weights without performing an exhaustive search. Using PACT and SAWB we realize a 2-bit QNN which achieves state of the art classification accuracy comparable to full precision networks while incurring no larger than $\mathcal{O}(n)$ computational overhead.

Fig. 1 shows that for AlexNet we not only achieve superior accuracy relative to other quantization techniques, but are also able to realize iso-accuracy relative to full-precision baseline with just 2-bit QNN. Overall, our 2-bit QNN achieves ($< 1\%$) accuracy loss for CIFAR10 tasks, and significantly improved accuracy relative to other quantization techniques for IMAGENET tasks. In comparison to other quantization schemes that increase the network size, this technique allows us to achieve iso-accuracy relative to full-precision models without necessitating this increase - thereby harnessing the full computational benefits of quantization.

The rest of the paper is organized as follows: Section 2 provides a summary of prior work on QNNs and challenges. We present a novel activation quantization scheme in Section 3 followed by a new weight quantization scheme in Section 4. In Section 5, we demonstrate the effectiveness of our quantization schemes using a 2-bit QNN across a set of popular CNNs.

## 2. Prior Work in QNN

There has been extensive research on quantizing weight and activation to minimize CNN computation and storage costs.

One of the earliest studies in weight quantization schemes (Hwang & Sung, 2014; Courbariaux et al., 2015) show that it is indeed possible to quantize weights to 1-bit (binary) or 2-bits (ternary), enabling an entire DNN model to fit effectively in resource-constrained platforms (e.g., mobile devices). Effectiveness of weight quantization techniques has been further improved (Li & Liu, 2016; Zhu et al., 2017), by ternarizing weights using statistical distribution of weight values or by tuning quantization scales during training.

To reduce the overhead of activations, prior work (Kim & Smaragdis, 2015; Hubara et al., 2016a; Rastegari et al., 2016) proposed the use of fully binarized neural networks where activations are quantized using 1-bit as well. More recently, activation quantization schemes using more general selections in bit-precision (Hubara et al., 2016b; Zhou et al., 2016) have been studied. However, these early techniques on weight and activation quantization show significant degradation in accuracy ($> 1\%$) for IMAGENET tasks (Russakovsky et al., 2015) when bit precision is reduced significantly ($\leq 2-bits$). To improve QNN accuracy for IMAGENET, more complex quantization schemes have been adopted. Weighted-entropy based quantization (WEQ, (Park et al., 2017)) uses iterative search algorithms for finding weight clusters or base/offset of logarithmic quantization for activation. Balanced quantization (BQ, (Zhou et al., 2017)) uses recursive partitioning of data into balanced bins. Half-wave Gaussian quantization (Cai et al., 2017) finds the quantization scale via Lloyd optimization on Normal distribution. These schemes demonstrated state-of-the-art QNN accuracy for neural networks for IMAGENET, such as Residual-Net, but they often involve computationally expensive sorting or recursive search during quantization.

More recently, alternative approaches to QNN focus on using simple quantization schemes but reduce quantization errors by performing more computations. One such approach is to increase the size of the network. Wide Residual Networks (Zagoruyko & Komodakis, 2016) demonstrates accuracy of a neural network can be improved by increasing the number of channels (called widening). Wide reduced-precision network (WRPN, (Asit Mishra, 2018b)) exploited this increased accuracy for QNN, by doubling the channel size to compensate for quantization error. (McDonnell, 2018) also employed Wide Residual Networks along with other modification in the training setting (e.g., warm-restart learning rate schedule) to improve accuracy for 1-bit weight quantization. Another way of taking advantage of extra computation in quantization is knowledge distillation (e.g., Apprentice, (Asit Mishra, 2018a)), where a teacher network (which is typically large and trained in full precision) is employed to help train the student network (one that is quantized). In all of these approaches accuracy improvement for QNN comes at the expense of large computational cost.

In summary, prior quantization techniques incur degradation in accuracy relative to full-precision, and use alterative approaches which requires significantly more computations to overcome the quantization errors. Our work is motivated by the desire to improve both weight and activation quantization schemes to achieve accuracy comparable to full-precision models while requiring no changes to the network structure.

## 3. Parameterized Clipping Activation

### 3.1. Challenge in Activation Quantization

Activation quantization becomes challenging when ReLU (the activation function most commonly used in CNNs) is used as the layer activation function. ReLU allows gradient of activations to propagate through deep layers and therefore achieves superior accuracy relative to other activation functions (Nair & Hinton, 2010). However, as the output of the ReLU function is unbounded, the quantization after ReLU requires a high dynamic range (i.e., more bit-precision). This is particularly problematic when the target bit-precision is low, e.g., 2-bits. In Fig.2 we present the training and validation errors of ResNet20 with the CIFAR10 dataset using ReLU and show that accuracy is significantly degraded when activation is quantized with ReLU.

It has been shown that this dynamic range problem can be alleviated by using a clipping activation function, which places an upper-bound on the output (Hubara et al., 2016b; Zhou et al., 2016). However, because of layer to layer and model to model differences - it is difficult to determine a globally optimal clipping value. In addition, as shown in Fig.2, even though the training error obtained using clipping with quantization is less than that obtained with quantized ReLU, the validation error is still noticeably higher than the baseline.

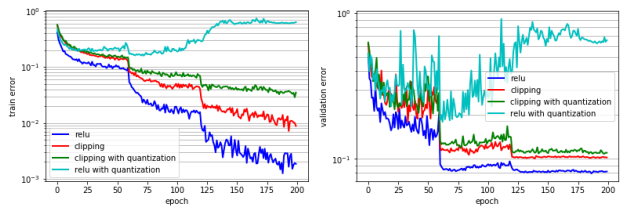

*Figure 2.* (a) Training error, (b) Validation error across epochs for different activation functions (relu and clipping) with and without quantization for the CIFAR10 ResNet20 model using the CIFAR10 dataset

### 3.2. Parameterized Clipping Activation

Building on these insights, we introduce PACT, a new activation quantization scheme in which the activation function has a parameterized clipping level, $\alpha$. $\alpha$ is dynamically adjusted via gradient descent-based training with the objective of minimizing the accuracy degradation arising from quantization. In PACT, the conventional ReLU activation function in CNNs is replaced with the following:

$$y = 0.5(|x| - |x - \alpha| + \alpha) = \begin{cases} 0, & x \in (-\infty, 0) \\ x, & x \in [0, \alpha) \\ \alpha, & x \in [\alpha, +\infty) \end{cases} \quad (1)$$

where $\alpha$ limits the range of activation to $[0, \alpha]$. This is illustrated in Fig.3(a). The truncated activation output is then linearly quantized to $k$ bits for the dot-product computations, where

$$y_q = round(y \cdot \frac{2^k - 1}{\alpha}) \cdot \frac{\alpha}{2^k - 1} \quad (2)$$

With this new activation function, $\alpha$ is a variable in the loss function, whose value can be optimized during training. For back-propagation, gradient $\frac{\partial y_q}{\partial \alpha}$ can be computed using the Straight-Through Estimator (STE) (Bengio et al., 2013) to estimate $\frac{\partial y_q}{\partial y}$ as 1. Thus,

$$\frac{\partial y_q}{\partial \alpha} = \frac{\partial y_q}{\partial y}\frac{\partial y}{\partial \alpha} = \begin{cases} 0, & x \in (-\infty, \alpha) \\ 1, & x \in [\alpha, +\infty) \end{cases} \quad (3)$$

The larger the $\alpha$, the more the parameterized clipping function resembles ReLU. To avoid large quantization errors due to a wide dynamic range, we include a L2-regularizer for $\alpha$ in the loss function. Fig.3(b) illustrates how the value of $\alpha$ changes during full-precision training of CIFAR10-ResNet20 starting with an initial value of 10 and using the L2-regularizer. It can be observed that $\alpha$ converges to values much smaller than the initial value as the training epochs proceed, thereby limiting the dynamic range of activations and minimizing quantization loss.

### 3.3. Analysis

#### 3.3.1. PACT IS AS EXPRESSIVE AS RELU

When used as an activation function of the neural network, PACT is as expressive as ReLU. This is because the clipping parameter, $\alpha$, introduced in PACT, allows flexibility in adjusting the dynamic range of activation for each layer, thus it can cover large dynamic range as needed. We demonstrate in the simple example below that PACT can reach the same solution as ReLU via SGD.

**Lemma 3.1.** *Consider a single-neuron network with PACT; $x = w \cdot a$, $y = PACT(x)$, where $a$ is input and $w$ is weight.*

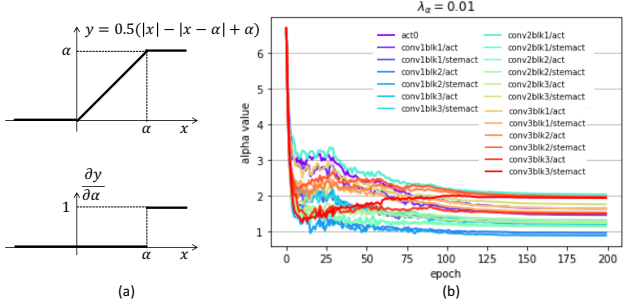

*Figure 3.* (a) PACT and its gradient. (b) Evolution of $\alpha$ values during training using a CIFAR10-ResNet20 model on the CIFAR10 dataset.

*This network can be trained with SGD to find the output the network with ReLU would produce.*

*Proof.* Consider a sample of training data $(a, y^*)$. For illustration purposes consider mean-square-error (MSE) as the cost function: $L = 0.5 \cdot (y^* - y)^2$.

If $x \leq \alpha$, then clearly the network with PACT behaves the same as the network with ReLU.

If $x > \alpha$, then $y = \alpha$ and $\frac{\partial y}{\partial \alpha} = 1$ from (1). Thus,

$$\frac{\partial L}{\partial \alpha} = \frac{\partial L}{\partial y} \cdot \frac{\partial y}{\partial \alpha} = \frac{\partial L}{\partial y} \quad (4)$$

Therefore, when $\alpha$ is updated by SGD,

$$\alpha_{new} = \alpha - \eta \frac{\partial L}{\partial \alpha} = \alpha - \eta \frac{\partial L}{\partial y} \quad (5)$$

where $\eta$ is a learning rate. Note that during this update, the weight is not updated as $\frac{\partial L}{\partial w} = \frac{\partial L}{\partial y} \cdot \frac{\partial y}{\partial x}(=0) \cdot a = 0$.

From MSE, $\frac{\partial L}{\partial y} = (y - y^*)$. Therefore, if $y^* > x$, $\alpha$ is increased for each update of (5) until $\alpha \geq x$, then the PACT network behaves the same as the ReLU network.

Interestingly, if $y^* \leq y$ or $y < y^* < x$, $\alpha$ is decreased or increased to converge to $y^*$. Note that in this case, ReLU would pass erroneous output $x$ to increase cost function, which needs to be fixed by updating $w$ with $\frac{\partial L}{\partial w}$. PACT, on the other hand, ignores this erroneous output by directly adapting the dynamic range to match the target output $y^*$. In this way, the PACT network can be trained to produce output which converges to the same target that the ReLU network would achieve via SGD.

□

In general cases, $\frac{\partial L}{\partial \alpha} = \sum_i \frac{\partial L}{\partial y_i}$, and PACT considers output of neurons together to change the dynamic range. There are two options: (1) if output $x_i$ is not clipped, then the network is trained via back-propagation of gradient to update weight, (2) if output $x_i$ is clipped, then $\alpha$ is increased or decreased based on how close the overall output is to the target. Hence, there are configurations under which SGD would lead to a solution close to the one which the network with ReLU would achieve. Fig. 4 demonstrates that CIFAR10-ResNet20 with PACT converges almost identical to the network with ReLU.

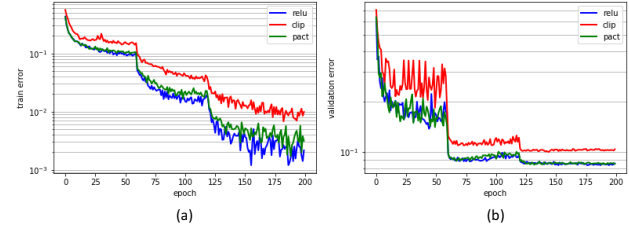

*Figure 4.* (a) Training error and (b) validation error of PACT for CIFAR10-ResNet20. Note that the convergence curve for PACT is almost identical to ReLU, although the dynamic range via trained clipping levels are much lower than ReLU.

### 3.3.2. BALANCING CLIPPING AND QUANTIZATION ERRORS

In Section 3.1, when we briefly discussed the challenges in activation quantization, we mentioned that there is a trade-off between errors due to clipping and quantization. As the clipping level increases, larger range of activation can be passed to the next layer of the neural network causing less clipping error ($ErrClip_i = max(x_i - \alpha, 0)$). However, the increased dynamic range incurs larger quantization error, since its magnitude is proportional to the clipping level ($ErrQuant_i \leq 0.5 \cdot \frac{\alpha}{2^k - 1}$, with $k$-bit quantization). This imposes the challenge of finding a proper clipping level to balance between clipping and quantization errors.

This trade-off can be better observed in Fig. 5a, which shows normalized mean-square-error caused by clipping and 2-bit quantization during training of the CIFAR10-ResNet20 with different clipping levels. It can be seen that activation functions with large dynamic range, such as ReLU, would suffer large quantization errors when the bit-precision is 2-bits. This explains why the network with ReLU fails to converge when the activation is quantized (Fig. 2).

PACT can find a balancing point between clipping and quantization errors. As explained in Section 3.3.1, PACT adjusts dynamic range based on how close the output is to the target. As both clipping and quantization errors distort output far from the target, PACT would increase or decrease the dynamic range during training to minimize both clipping and quantization errors.

Fig. 5b shows how PACT balances the clipping and quantization errors during training. CIFAR10-ResNet20 is trained with clipping activation function with varying clipping level $\alpha$ from 1 to 16. When activation is quantized with 2-bit, the network trained with clipping activation shows significant accuracy degradation as $\alpha$ increases. This is consistent with the trend in quantization error we observed in Fig. 5a. In this case, PACT achieves the best accuracy one of the clipping activation could achieve, but without exhaustively sweeping over different clipping levels. In other words, PACT auto-tunes the clipping level to achieve best accuracy without incurring significant computation overhead. PACT's auto-tuning of dynamic range is critical in efficient yet robust training of large scale quantized neural networks, especially because it does not increase the burden for hyper-parameter tuning. In fact, we used the same hyper-parameters as well as the original network structure for all the models we tested, except replacing ReLU to PACT, when we applied activation quantization.

Without quantization, there is a trend that validation error is small when $\alpha$ is large. Surprisingly, some of the cases even outperforms the ReLU network. In this case, PACT also achieves comparable accuracy as ReLU, confirming its expressivity discussed in Section 3.3.1.

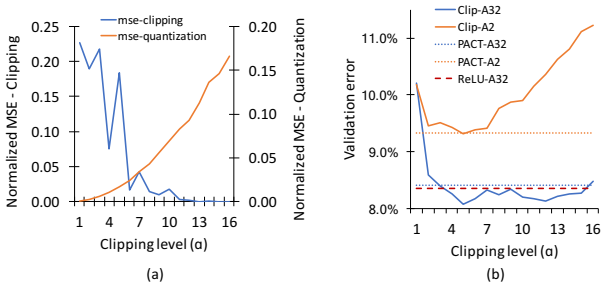

*Figure 5.* Experiment on CIFAR10-ResNet20 to validate that PACT balances clipping and quantization errors. (a) Trade-off between clipping and quantization error. (b) PACT achieving lowest validation error that clipping activation can achieve without exhaustive search over clipping level $\alpha$.

## 4. Statistics-Aware Weight Binning

In addition to the activation quantization, we also propose a statistical-aware weight binning (SAWB) scheme for the weight quantization that aims to minimize the mean-square error between the quantized and the full-precision weight distribution. We will focus on 2-bit evenly-spaced weight quantization, though we will also show that SAWB is applicable to a wide range of quantization levels ((i.e. $n_{bin} = 2, 3, 4, 8, 16, 32$, corresponding to binarization, ternarization, 2-5 bit quantization, respectively)

Fig. 6 illustrates a 2-bit (4 bins) evenly-spaced SAWB scheme. We choose symmetrical and uniformly distributed quantization bins because it allows for a hardware friendly multiplication and accumulation (MAC) design. For SAWB, the scaling factor, $\alpha_w$, defines the largest quantization level and each weight is quantized to the nearest bin. For each weight distribution, there exists an optimal $\alpha_w^*$ such that the MSE error is minimized. That is,

$$\alpha_w^* = \arg\min_{\alpha_w} ||w - w_q||^2 \quad (6)$$

The key here is to find $\hat{\alpha}_w^* \approx \alpha_w^*$ efficiently (i.e., without an exhaustive search). SAWB can quickly determine $\hat{\alpha}_w^*$ for a variety of weight distribution based on the distribution's statistic (i.e., $E(w^2)$ and $E(|w|)$) and results in a $\mathcal{O}(n)$ complexity, where $n$ is the number of weight.

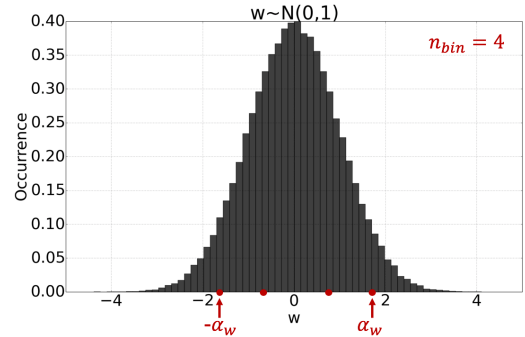

*Figure 6.* Quantization scheme for weights.

To begin, we first determine the optimal $\alpha_w^*$ of the following six distributions for a wide range of quantization levels : Gaussian distribution with [$\mu$:0,$\sigma^2$:1]; Uniform distribution with $\mu$=0 and a range of [-1:1],Laplace distribution with $\mu$=0 and decay of 1; logistic distribution with $\mu$=0 and scale 1; triangle distribution with $\mu$=0 and extreme 2; and von Mises distribution with center 0 and dispersion 4.

In Fig. 7, we plot $\frac{\alpha_w^*}{E(|w|)}$ (x axis) versus $\frac{\sqrt{E(w^2)}}{E(|w|)}$ (y axis) for the six distributions under consideration (points in different colors) with different $n_{bin}$ values. Given a bin level $n_{bin}$, linear regression can be applied to derive $\alpha_w^*$ as a function of $\sqrt{E(w^2)}$ and $E(|w|)$. As a result, during each mini-batch of training, we first compute $E(|w|)$ and $E(w^2)$, and then applied the linear coefficients that we empirically determined to calculate $\hat{\alpha}_w^*$, which is then used for the weight quantization.

Table 1 summarizes the square error (SE) of layer 11 and layer 15's weights based on the optimal and SAWB estimated scaling factor in a ResNet20 network with CIFAR10 dataset employing a 2-bit evenly-spaced weight quantization scheme. $\alpha_w^*$ is determined through an off-line exhaustive search. It is evident that the $\hat{\alpha}_w^*$ determined by the proposed SAWB incurs less than 7% additional SE at anytime during the training. As training progresses, the SE error between the optimal and SAWB case increases, suggesting that the weight distribution gradually diverges from the initial Gaussian-like distribution. An interesting future work to mitigate this problem is to calibrate the linear coefficients every fixed interval of epoch by including the statistical information of the weight distribution in each layer to achieve an improved correlation.

Table 1. Square Error (SE) of the optimal and SAWB estimated scaling factor of CIFAR10-ResNet20 at different epoch with 2-bit weight quantization

| Epoch | 1 | 40 | 80 | 120 | 160 | 200 |
|---|---|---|---|---|---|---|
| Layer 11 Weight | | | | | | |
| Optimal SE | 6.77 | 12.21 | 10.45 | 8.24 | 7.92 | 7.75 |
| SAWB SE | 6.94 | 12.43 | 10.64 | 8.40 | 8.13 | 7.98 |
| Layer 15 Weight | | | | | | |
| Optimal SE | 13.72 | 48.35 | 40.49 | 31.24 | 29.96 | 29.22 |
| SAWB SE | 13.73 | 50.24 | 42.38 | 32.79 | 31.76 | 31.11 |

Table 2. Accuracy of CIFAR10-ResNet20/32/44/56 for quantizing activation only, weight only, and both weight and activation. (d,p,s) indicates (DoReFa,PACT,SAWB), respectively. "fpsc" indicates full-precision short-cut. PACT-SAWB achieves accuracy $\leq 0.5\%$ for individual quantization, and $\leq 1\%$ for quantizing both weight and activation, compared to full precision baseline. In comparison DoReFa net achieves an error-rate of 11.79% when using 2-bit weights and activations - resulting in a > 3% degradation

| Layers | 20 | 32 | 44 | 56 |
|---|---|---|---|---|
| Full-Precision (32-bit) | 8.49% | 7.49% | 6.84% | 6.77% |
| W32-Ad2 | 9.92% | 9.27% | 8.89% | 8.48% |
| W32-Ad2-fpsc | 9.31% | 8.60% | 7.95% | 8.03% |
| W32-Ap2 | 9.51% | 8.44% | 8.02% | 7.76% |
| **W32-Ap2-fpsc** | **8.64%** | **7.72%** | **7.29%** | **7.01%** |
| Wd2-A32 | 9.14% | 8.51% | 7.90% | 7.40% |
| Wd2-A32-fpsc | 9.23% | 8.44% | 7.97% | 7.41% |
| Ws2-A32 | 9.27% | 8.80% | 7.84% | 7.45% |
| **Ws2-A32-fpsc** | **9.08%** | **7.74%** | **7.39%** | **7.07%** |
| Ws2-Ap2 | 10.77% | 9.57% | 9.39% | 8.76% |
| **Ws2-Ap2-fpsc** | **9.35%** | **8.36%** | **7.61%** | **7.48%** |

## 5. Experiments

We implemented PACT and SAWB in Tensorflow (Abadi et al., 2015) using Tensorpack (Zhou et al., 2016). To demonstrate the effectiveness of PACT and SAWB, we studied several well-known CNNs: ResNet20/32/44/56 (He et al., 2016b) for the CIFAR10 dataset, and AlexNet and ResNet18/50 models for the IMAGENET dataset. Implementation detail for each network is described in the Appendix A. Note that the baseline networks use the same hyper-parameters and ReLU activation functions as described in the references. For PACT experiments, we only replace ReLU with PACT but otherwise use the same hyper-parameters. The networks are trained from scratch and the first/last layers are not quantized during training.

### 5.1. Full-Precision Shortcut for Improving Quantization Accuracy for ResNet

We made a key observation that applying quantization naively to the shortcut (residual) path impacts accuracy because the shortcut path plays a key role in backpropagating gradient information throughout the layers and quantization fundamentally limits the flow of gradients. It is also mentioned in (He et al., 2016a) that keeping the shortcut connections more direct is beneficial for ease of optimization and to reduce overfitting. In the same vein, we did not quantize the input activations or the weights in the residual path. Our quantization results show that when using full-precision only for shortcut path the accuracy of all the tested quantized ResNets is within $\leq 1\%$ of the full-precision ResNets. Note that the shortcut in ResNet takes a negligible portion of the overall compute, e.g., <1% for IMAGENET-ResNet18.

### 5.2. Quantization Accuracy for PACT and SAWB

We first evaluate our activation and weight quantization schemes using CIFAR10-ResNet with (20,32,44,56) layers. Table 2 summarizes the accuracy for quantizing activation only, weight only, and both weights and activations. DoReFa's quantization results are also shown for comparison. For activation only quantization, PACT achieves superior accuracy for all the cases relative to DoReFa. Full-precision shortcut (postfix -fpsc) further improves accuracy by $\leq 1\%$ for both of them. Overall, PACT with full-precision shortcut achieves accuracy within $0.5\%$ of the full-precision accuracy. In the case of weight quantization, full-precision short-cut is consistently more helpful for SWAB than DoReFa. Overall, SAWB with full-precision short-cut also achieves accuracy within $5\%$ of the full-precision baseline, outperforming DoReFa. Putting it all together, PACT-SAWB achieves accuracy within $1\%$ of the full precision accuracy.

### 5.3. Comparison of Quantization Accuracy with Other QNNs

We implemented DoReFA (Zhou et al., 2016) from the same setting (i.e., they share the same baseline). For further comparison, we have used the accuracy results reported in the following prior work: BQ (Zhou et al., 2017), WEQ (Park et al., 2017), WRPN (Asit Mishra, 2018b), Apprentice (Asit Mishra, 2018a). We compared the accuracy results from AlexNet as well as ResNet18 and ResNet50. Table 3 summarizes quantization accuracy for the all prior work and our work. Detailed experimental setting for each of these papers can be found in Appendix B.

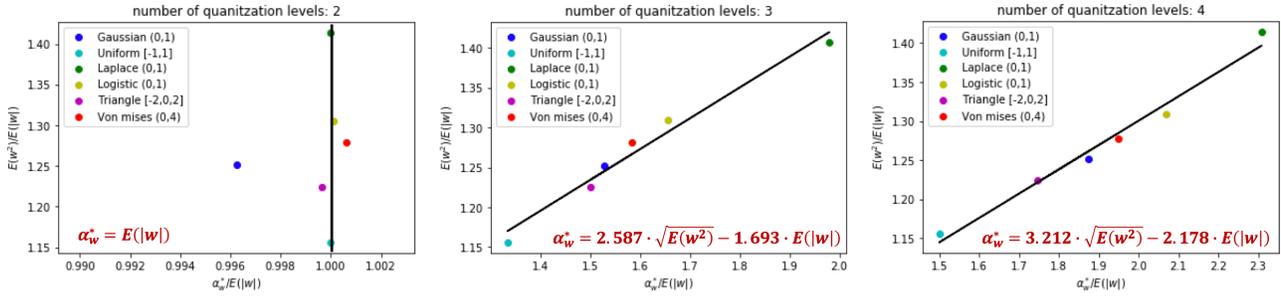

*Figure 7.* Derivation of empirical equation for determining optimal quantization from statistical distribution of weights.

*Table 3.* Comparison of quantization accuracy (top-1) with other QNNs. Full-precision as well as quantization accuracy with different bit-precision are presented (WX-AY indicates X-bit weight and Y-bit activation quantization). Our quantization schemes achieve iso-accuracy for AlexNet. Furthermore, our quantization schemes achieved state of the art accuracy for ResNet18 and ResNet50 outperforming all the other QNNs compared.

| Alexnet | Ours (PACT-SAWB) | | DoReFa | WRPN | | | | WRPN-x2 | | WEQ | | | | BQ | |
|---|---|---|---|---|---|---|---|---|---|---|---|---|---|---|---|
| | Full-Prec | W2-A2 | W2-A2 | Full-Prec | W2-A2 | W2-A4 | W4-A4 | W2-A2 | W2-A4 | Full-Prec | W2-A2 | W2-A4 | W4-A4 | Full-Prec | W2-A2 |
| Accuracy | 55.1% | 55.0% | 54.1% | 57.2% | 51.3% | 50.5% | 54.4% | 55.8% | 57.2% | 57.1% | 50.6% | 54.4% | 55.9% | 57.1% | 55.7% |
| Degradation | | 0.1% | 1.0% | | 5.9% | 6.7% | 2.8% | 1.4% | 0.0% | | 6.5% | 2.7% | 1.2% | | 1.4% |

| ResNet18 | Ours (PACT-SAWB) | | DoReFa | BQ | | |
|---|---|---|---|---|---|---|
| | Full-Prec | W2-A2-fpsc | W2-A2 | Full-Prec | W2-A2 | W32-A2 |
| Accuracy | 70.4% | 67.0% | 62.6% | 68.2% | 59.4% | 62.1% |
| Degradation | | 3.4% | 7.7% | | 8.8% | 6.1% |

| ResNet50 | Ours (PACT-SAWB) | | DoReFa | Apprentice | |
|---|---|---|---|---|---|
| | Full-Prec | W2-A2-fpsc | W2-A2 | Full-Prec | W2-A8 |
| Accuracy | 76.9% | 74.2% | 67.1% | 76.2% | 72.8% |
| Degradation | | 2.7% | 9.8% | | 3.4% |

From Table 3 we observe that for AlexNet, PACT-SAWB with 2-bit QNN achieve negligible degradation compared to full-precision baseline. In contrast, for the same size of the neural network, all the prior quantization schemes (DoReFa, WRPN, WEQ, BQ) suffer more than 1% degradation with 2-bit weight and activation.

WRPN-x2 which widens the number of channels by 2x has 1.4% accuracy degradation when 2-bit weight and 2-bit activation are used. WRPN-x2 requires two more activation bits to achieve baseline accuracy. However, increasing network size increases the total computations and impacts the latency of classification. For example, Fig. 8 shows the normalized energy consumption of reduced-precision processing unit (Multiply-Accumulate unit, MAC), which is implemented in 14nm CMOS. In the case of WRPN, there is $9\times$ energy saving compared to 16-bit MAC thanks to reduced precision. But WRPN-2x sacrifices this potential gain because widening of a network by $2\times$ results in $4\times$ more MAC computations (shown as $4\times$ gap in Fig. 8 relative to WRPN). So the additional computation cost makes the choice of increasing the network size not an attractive option for improving quantization accuracy.

### 5.4. Bridging Accuracy Gap for Higher QNN Performance

Since increasing the channel size to improve quantization accuracy is complementary to PACT-SAWB, we also studied the impact on quantization accuracy by increasing the number of channels as presented in WRPN. For compari-son, we also applied DoReFa's quantization scheme, since WRPN's quantization scheme is a simpler version of this (He et al., 2016a). CIFAR10-ResNet20 is used for the experiment, in which the 1st/2nd/3rd blocks have 16/32/64 channels, respectively.

Fig. 9 shows that as we increase the number of channels from 1x to 2x shown by "channel increasing factor" in the X-axis, the validation error is reduced. We observe that to achieve baseline full-precision accuracy, DoReFa's quantization scheme requires 2x more channels relative to PACT-SAWB. PACT-SAWB outperfoms baseline full-precision accuracy with $1.25\times$ channels confirming that starting with a more robust quantization scheme enables bridging accuracy gap with less additional computation overhead.

## 6. Conclusion

In this paper, we propose novel techniques that target weight and activation quantizations separately resulting in an overall quantized neural network (QNN). The activation quantization technique, PArameterized Clipping acTivation (PACT), uses an activation clipping parameter $\alpha$ that is optimized during training to find the right quantization scale. The weight quantization scheme, statistics-aware weight binning (SAWB), finds the optimal scaling factor that minimizes the quantization error based on the statistical characteristics of the distribution of weights without performing an exhaustive search. Our evaluations show that our quantization scheme achieves best overall accuracy relative to other prior quanti-zation techniques. We demonstrate that using a 2-bit QNN, we achieve iso-accuracy relative to full-precision baseline for AlexNet, ($< 1\%$) accuracy loss for CIFAR10 tasks, and best accuracy relative to other quantization techniques for IMAGENET tasks. Furthermore, when we also explore the complementary technique of increasing the network size to reduce the quantization error in our QNN, our results show that we are able to achieve iso-accuracy relative to full-precision with significantly lower increase in the model size relative to other quantization techniques. Overall, our results show that using PACT and SAWB we realize a 2-bit QNN which achieves classification accuracy comparable to full precision networks while incurring no larger than $\mathcal{O}(n)$ computational overhead.

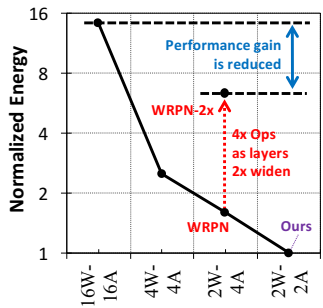

*Figure 8.* Impact of increasing network size to performance. Since WRPN-2x increases the number of channels by 2x, the total number of compute operations increases by 4x, reducing the potential energy savings achievable from reduced-bit computation of QNN.

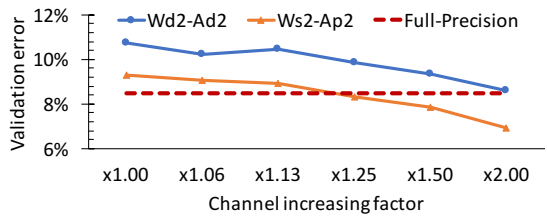

*Figure 9.* Impact of quantization accuracy on performance. DoReFa-Net and our quantization schemes are used in CIFAR10-ResNet20 training, where the number of channels is increased up to 2x to compensate accuracy loss, as done in WRPN. Note that our scheme requires much lower increase to achieve the baseline accuracy.

## Acknowledgments

The authors would like to thank Naigang Wang, Daniel Brand, Chia-Yu Chen, Ankur Agrawal and I-Hsin Chung for helpful discussions and supports. This research was supported by IBM Research AI, IBM SoftLayer, and IBM Cognitive Computing Cluster (CCC).

## References

Martín Abadi, Ashish Agarwal, Paul Barham, Eugene Brevdo, Zhifeng Chen, Craig Citro, Greg S. Corrado, Andy Davis, Jeffrey Dean, Matthieu Devin, Sanjay Ghemawat, Ian Goodfellow, Andrew Harp, Geoffrey Irving, Michael Isard, Yangqing Jia, Rafal Jozefowicz, Lukasz Kaiser, Manjunath Kudlur, Josh Levenberg, Dan Mané, Rajat Monga, Sherry Moore, Derek Murray, Chris Olah, Mike Schuster, Jonathon Shlens, Benoit Steiner, Ilya Sutskever, Kunal Talwar, Paul Tucker, Vincent Vanhoucke, Vijay Vasudevan, Fernanda Viégas, Oriol Vinyals, Pete Warden, Martin Wattenberg, Martin Wicke, Yuan Yu, and Xiaoqiang Zheng. TensorFlow: Large-Scale Machine Learning on Heterogeneous Systems, 2015. URL https://www.tensorflow.org/. Software available from tensorflow.org.

Dan Alistarh Antonio Polino, Razvan Pascanu. Model compression via distillation and quantization. *ICLR*, 2018.

Debbie Marr Asit Mishra. Apprentice: Using Knowledge Distillation Techniques To Improve Low-Precision Network Accuracy. *International Conference on Learning Representations*, 2018a.

Jeffrey J Cook Debbie Marr Asit Mishra, Eriko Nurvitadhi. WRPN: Wide Reduced-Precision Networks. *ICLR*, 2018b.

Yoshua Bengio, Nicholas Léonard, and Aaron C. Courville. Estimating or Propagating Gradients Through Stochastic Neurons for Conditional Computation. *CoRR*, abs/1308.3432, 2013.

Zhaowei Cai, Xiaodong He, Jian Sun, and Nuno Vasconcelos. Deep Learning With Low Precision by Half-Wave Gaussian Quantization. *IEEE Conference on Computer Vision and Pattern Recognition (CVPR)*, July 2017.

Matthieu Courbariaux, Yoshua Bengio, and Jean-Pierre David. BinaryConnect: Training Deep Neural Networks with binary weights during propagations. *CoRR*, abs/1511.00363, 2015.

Jonas Gehring, Michael Auli, David Grangier, Denis Yarats, and Yann N. Dauphin. Convolutional Sequence to Sequence Learning. *CoRR*, abs/1705.03122, 2017.

Kaiming He, Xiangyu Zhang, Shaoqing Ren, and Jian Sun. Identity Mappings in Deep Residual Networks. *CoRR*, abs/1603.05027, 2016a.

Kaiming He, Xiangyu Zhang, Shaoqing Ren, and Jian Sun. Deep Residual Learning for Image Recognition. *IEEE Conference on Computer Vision and Pattern Recognition (CVPR)*, pp. 770–778, 2016b.

Bridging the Accuracy Gap for 2-bit Quantized Neural Networks (QNN)

Itay Hubara, Matthieu Courbariaux, Daniel Soudry, Ran El-Yaniv, and Yoshua Bengio. Binarized Neural Networks. *NIPs*, pp. 4107–4115, 2016a.

Itay Hubara, Matthieu Courbariaux, Daniel Soudry, Ran El-Yaniv, and Yoshua Bengio. Quantized Neural Networks: Training Neural Networks with Low Precision Weights and Activations. *CoRR*, abs/1609.07061, 2016b.

Kyuyeon Hwang and Wonyong Sung. Fixed-point Feedforward Deep Neural Network Design Using Weights +1, 0, and -1. In *IEEE Workshop on Signal Processing Systems (SiPS)*, pp. 1–6, Oct. 2014.

Anke Schmeink Tim Genewein Jan Achterhold, Jan Mathias Koehler. Variational Network Quantization. *ICLR*, 2018.

Minje Kim and Paris Smaragdis. Bitwise Neural Networks. *ICML Workshop on Resource-Efficient Machine Learning*, 2015.

Alex Krizhevsky and G Hinton. Convolutional deep belief networks on cifar-10. *Unpublished manuscript*, 40, 2010.

Alex Krizhevsky, Ilya Sutskever, and Geoffrey E. Hinton. ImageNet Classification with Deep Convolutional Neural Networks. In *Advances in Neural Information Processing Systems 25 (NIPS)*, pp. 1097–1105, 2012.

Fengfu Li and Bin Liu. Ternary Weight Networks. *CoRR*, abs/1605.04711, 2016.

James T. Kwok Lu Hou. Loss-aware Weight Quantization of Deep Networks. *International Conference on Learning Representations*, 2018.

Mark D. McDonnell. Training wide residual networks for deployment using a single bit for each weight. *ICLR*, 2018.

Vinod Nair and Geoffrey E. Hinton. Rectified Linear Units Improve Restricted Boltzmann Machines. *27th International Conference on Machine Learning (ICML)*, pp. 807–814, 2010.

Eunhyeok Park, Junwhan Ahn, and Sungjoo Yoo. Weighted-Entropy-Based Quantization for Deep Neural Networks. *IEEE Conference on Computer Vision and Pattern Recognition (CVPR)*, July 2017.

Mohammad Rastegari, Vicente Ordonez, Joseph Redmon, and Ali Farhadi. XNOR-Net: ImageNet Classification Using Binary Convolutional Neural Networks. *CoRR*, abs/1603.05279, 2016.

Olga Russakovsky, Jia Deng, Hao Su, Jonathan Krause, Sanjeev Satheesh, Sean Ma, Zhiheng Huang, Andrej Karpathy, Aditya Khosla, Michael Bernstein, Alexander C. Berg, and Li Fei-Fei. ImageNet Large Scale Visual Recognition Challenge. *International Journal of Computer Vision (IJCV)*, 115(3):211–252, 2015.

Sergey Zagoruyko and Nikos Komodakis. Wide Residual Networks. *arXiv preprint arXiv:1605.07146*, 2016.

Ying Zhang, Mohammad Pezeshki, Philemon Brakel, Saizheng Zhang, César Laurent, Yoshua Bengio, and Aaron C. Courville. Towards End-to-End Speech Recognition with Deep Convolutional Neural Networks. *CoRR*, abs/1701.02720, 2017.

Shuchang Zhou, Zekun Ni, Xinyu Zhou, He Wen, Yuxin Wu, and Yuheng Zou. DoReFa-Net: Training Low Bitwidth Convolutional Neural Networks with Low Bitwidth Gradients. *CoRR*, abs/1606.06160, 2016.

Shuchang Zhou, Yuzhi Wang, He Wen, Qinyao He, and Yuheng Zou. Balanced Quantization: An Effective and Efficient Approach to Quantized Neural Networks. *CoRR*, abs/1706.07145, 2017.

Chenzhuo Zhu, Song Han, Huizi Mao, and William J. Dally. Trained Ternary Quantization. *International Conference on Learning Representations (ICLR)*, 2017.

## A. CNN Implementation Details

In this section, we summarize details of our CNN implementation as well as our training settings, which is based on the default networks provided by Tensorpack (Zhou et al., 2016). Unless mentioned otherwise, ReLU following BatchNorm is used for activation function of the convolution (CONV) layers, and Softmax is used for the fully-connected (FC) layer. Note that the baseline networks use the same hyper-parameters and ReLU activation functions as described in the references. For PACT experiments, we only replace ReLU into PACT but the same hyper-parameters are used. All the time the networks are trained from scratch.

The CIFAR10 dataset (Krizhevsky & Hinton, 2010) is an image classification benchmark containing $32 \times 32$ pixel RGB images. It consists of 50K training and 10K test image sets. We used the "full pre-activation" ResNet structure (He et al., 2016a) which consists of a CONV layer followed by 3 ResNet blocks (18/30/42/54 CONV layers with 3x3 filter, depending on choice) and a final FC layer. We used stochastic gradient descent (SGD) with momentum of 0.9 and learning rate starting from 0.1 and scaled by 0.1 at epoch 60, 120. L2-regularizer with decay of 0.0002 is applied to weight. The mini-batch size of 128 is used, and the maximum number of epochs is 200.

The IMAGENET dataset (Russakovsky et al., 2015) consists of 1000-categories of objects with over 1.2M training and 50K validation images. Images are first resized to 256 256 and randomly cropped to 224224 prior to being used as input to the network. We used a modified AlexNet, ResNet18 and ResNet50.

We used AlexNet network (Krizhevsky et al., 2012) in which local contrast renormalization (R-Norm) layer is replaced with BatchNorm layer. We used ADAM with epsilon $10^{-5}$ and learning rate starting from $10^{-4}$ and scaled by 0.2 at epoch 56 and 64. L2-regularizer with decay factor of $5 \times 10^{-6}$ is applied to weight. The mini-batch size of 128 is used, and the maximum number of epochs is 100.

ResNet18 consists of a CONV layer followed by 8 ResNet blocks (16 CONV layers with 3x3 filter) and a final FC layer. "full pre-activation" ResNet structure (He et al., 2016a) is employed. ResNet50 consists of a CONV layer followed by 16 ResNet "bottleneck" blocks (total 48 CONV layers) and a final FC layer. "full pre-activation" ResNet structure (He et al., 2016a) is employed.

For both ResNet18 and ResNet50, we used stochastic gradient descent (SGD) with momentum of 0.9 and learning rate starting from 0.1 and scaled by 0.1 at epoch 30, 60, 85, 95. L2-regularizer with decay of $10^{-4}$ is applied to weight. The mini-batch size of 256 is used, and the maximum number of epochs is 110.

## B. Comparison with Related Work

- DoReFa-Net (DoReFa, (Zhou et al., 2016)): A general bit-precision uniform quantization schemes for weight, activation, and gradient of DNN training. We compared the experimental results of DoReFa for CIFAR10, AlexNet and ResNet18 under the same experimental setting as PACT.

- Balanced Quantization (BQ, (Zhou et al., 2017)): A quantization scheme based on recursive partitioning of data into balanced bins. We compared the reported top-1/top-5 validation accuracy of their quantization scheme for AlexNet and ResNet18.

- Quantization using Wide Reduced-Precision Networks (WRPN, (Asit Mishra, 2018b)): A scheme to increase the number of filter maps to increase robustness for activation quantization. We compared the reported top-1 accuracy of their quantization with various weight/activation bit-precision for AlexNet.

- Weighted-entropy-based quantization (WEQ, (Park et al., 2017)): A quantization scheme that considers statistics of weight/activation. We compared the top-1 reported accuracy of their quantization with various bit-precision for AlexNet, where the first and last layers are not quantized.

- Apprentice (Apprentice, (Asit Mishra, 2018a)): A quantization scheme based on knowledge distillation where the teacher network (same size or larger network, trained in full-precision) helps the student network (small and quantized) to learn parameters to tolerate quantization errors. We compared the top-1 reported accuracy of their 2-bit weight and 8-bit activation quantization for ResNet50 (with ResNet-101 as the teacher network). The first and last layers are not quantized.